\documentclass[10pt,letterpaper]{article}

\usepackage{cogsci}
\usepackage{pslatex}
\usepackage[natbibapa]{apacite}

\usepackage{amsmath}
\usepackage{graphicx}
\usepackage{multirow}
\usepackage{booktabs}
\usepackage{textcomp}
\usepackage{tabularx}
\usepackage{tabulary}
\usepackage{array}

\title{Evaluating vector-space models of analogy}

 \author{
{\large \bf Dawn Chen (sdawnchen@gmail.com)} \\
{\large \bf Joshua C. Peterson (peterson.c.joshua@gmail.com)}\\
{\large \bf Thomas L. Griffiths (tom\_griffiths@berkeley.edu)} \\
  Department of Psychology, University of California, Berkeley, CA 94720 USA}

\begin{document}
\maketitle

\begin{abstract}

Vector-space representations provide geometric tools for reasoning about the similarity of a set of objects and their relationships. Recent machine learning methods for deriving vector-space embeddings of words (e.g., \texttt{word2vec}) have achieved considerable success in natural language processing. These vector spaces have also been shown to exhibit a surprising capacity to capture verbal analogies, with similar results for natural images, giving new life to a classic model of analogies as parallelograms that was first proposed by cognitive scientists. We evaluate the parallelogram model of analogy as applied to modern word embeddings, providing a detailed analysis of the extent to which this approach captures human relational similarity judgments in a large benchmark dataset. We find that that some semantic relationships are better captured than others. We then provide evidence for deeper limitations of the parallelogram model based on the intrinsic geometric constraints of vector spaces, paralleling classic results for first-order similarity.

\textbf{Keywords:}
analogy; \texttt{word2vec}; \texttt{GloVe}; vector space models
\end{abstract}


\section{Introduction}

Recognizing that two situations have similar patterns of \textit{relationships}, even though they may be superficially dissimilar, is essential for intelligence. This ability allows a reasoner to transfer knowledge from familiar situations to unfamiliar but analogous situations, enabling analogy to become a powerful teaching tool in math, science, and other fields \citep{richland2015analogy}. Computational modeling of analogy has primarily focused on comparing structured representations that contain labeled relationships between entities \citep{gentner2011computational}. However, the questions of where these relations come from and how to determine that the relationship between one pair of entities is the same as that between another pair are an unsolved mystery in such models. Some models, such as DORA \citep{doumas2008theory} and BART \citep{lu2012bayesian}, try to learn relations from examples, but have only demonstrated success on comparative relations such as \textit{larger}.

\begin{figure}[!ht]
   \centering
    \vspace{-4mm}
  \includegraphics[width=0.45\linewidth,keepaspectratio]{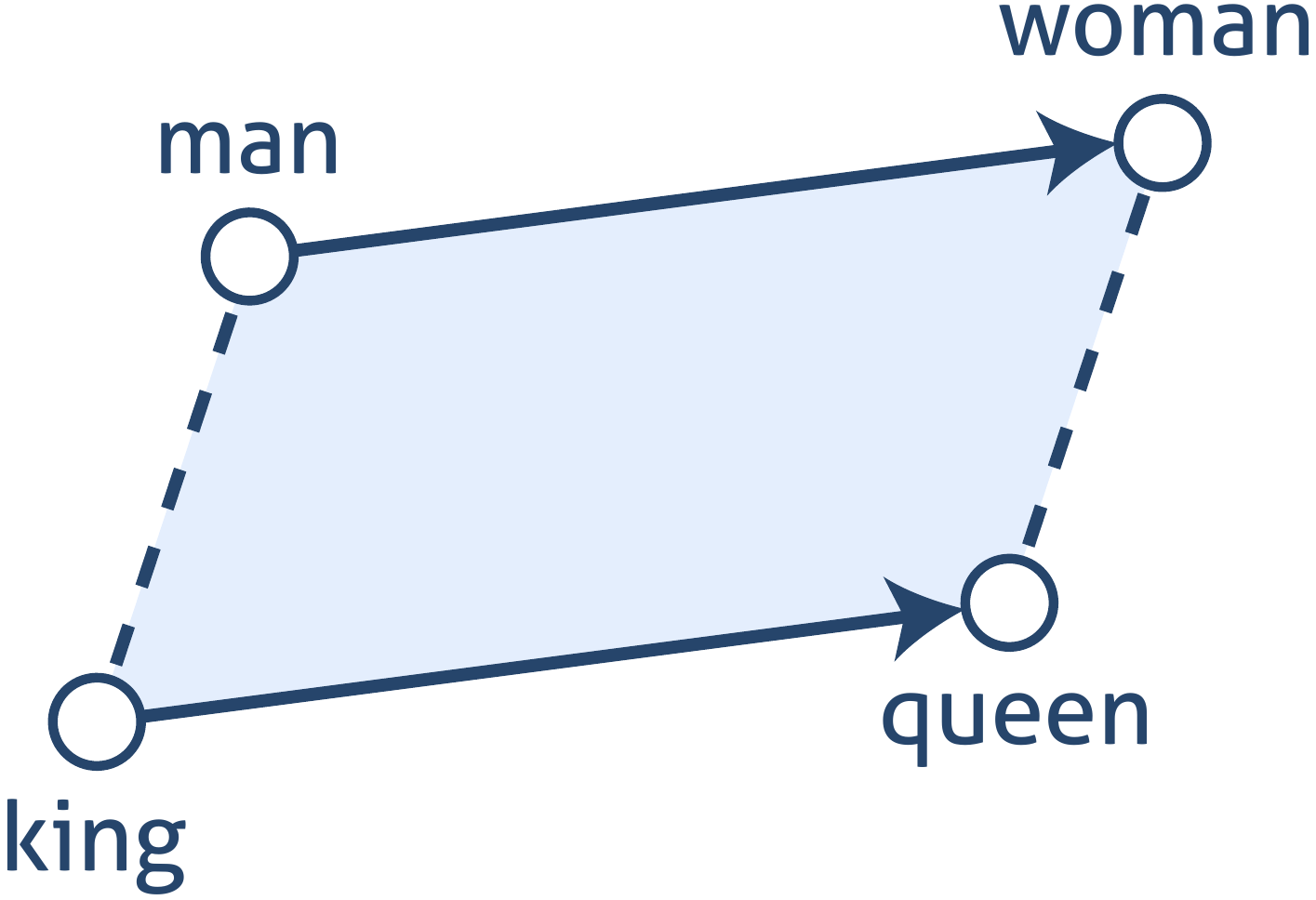}
    \vspace{-4mm}
    \caption{The parallelogram model of analogy completes the analogy {\em king} : \ {\em queen} :: {\em man} : ? by adding the difference vector between {\em king} and {\em queen} to {\em man}. This forms a parallelogram in the underlying vector space.}
    \label{parallelogram}
\end{figure}

Another possibility is that the representations of entities themselves contain the information necessary to infer relationships between entities and that relations do not need to be learned separately. An instantiation of this hypothesis is the parallelogram model of analogy (see Figure \ref{parallelogram}), first proposed by \cite{Rumelhart1973} over 40 years ago. In this model, entities are represented as points in a Euclidean space and relations between entities are represented as their difference vectors. Even though two pairs of points may be far apart in the space (i.e., they are featurally dissimilar), they are considered relationally similar as long as their difference vectors are similar. Although \citeauthor{Rumelhart1973} found that this simple model worked well for a small domain of animal words with vectors obtained using multidimensional scaling, little progress was made on the parallelogram model after the initial proposal, with the exception of a handful of reasonably successful applications \cite[see][]{ehresman1978perception}.


However, in the past few years, the parallelogram model was reincarnated in the machine learning literature through popular word embedding methods such as \texttt{word2vec} \citep{Mikolov2013-word2vec1} and \texttt{GloVe} \citep{pennington2014glove}. These word representations enable verbal analogy problems such as \textit{king} : \textit{queen} :: \textit{man} : ? to be solved through simple vector arithmetic, i.e., $v_{\textit{queen}} - v_{\textit{king}} + v_{\textit{man}}$ results in a vector very close (in terms of cosine distance) to $v_{\textit{woman}}$. Word embeddings like \texttt{word2vec} and \texttt{GloVe} have also been used successfully in a variety of other natural language processing tasks, suggesting that these representations may indeed contain enough information for relations to be inferred from them directly. Recently, researchers in computer vision have been successful in extracting feature spaces that exhibit similar properties in both explicit (supervised) \citep{radford2015unsupervised} and implicit (unsupervised) \citep{reed2015deep} ways, yielding linearized semantic image transformations such as object rotations and high-level human face interpolations. The potential for applying the parallelogram model of analogy to vector space models appears to be domain-agnostic, broadly applicable to both semantic and perceptual domains. This suggests a promising cognitive model and provides the opportunity to evaluate a classic theory in large-scale, ecologically valid contexts.


In this paper, we evaluate the parallelogram model of analogy as applied to modern vector-space representations of words. Focusing on the predictions that this approach makes about the relational similarity of words, we provide a new dataset of over 5,000 comparisons between word pairs that exemplify 10 different types of semantic relations. We find that the parallelogram model captures human relational similarity judgments for some semantic relations, but not others. We then show that human relational similarity judgments for pairs of words violate the geometric constraints of symmetry and triangle inequality, just as \cite{tversky1977features} showed for judgments of first-order similarity between words. This creates a challenge for \textit{any} vector space model that aims to make predictions about relational similarity.


\section{Relational Similarity}

One way to evaluate vector space models such as \texttt{word2vec} and \texttt{GloVe} as accounts of analogy is to compare their assessments of relational similarity -- the similarity of the relation between one pair of words to that of another -- with human judgments. A good foundation for this task is the SemEval-2012 Task 2 dataset \citep{Jurgens2012}, which contains prototypicality scores based on human data for word pairs that exemplify 79 different semantic relations. These relations were taken from a taxonomy of semantic relations \citep{Bejar1991} and are subtypes of 10 general types, such as \textsc{class-inclusion}, \textsc{similar}, and \textsc{contrast}. Participants were given three or four paradigmatic examples of a relation and asked to generate additional examples of the same relation. A total of 3,218 unique word pairs were generated for the 79 relations, with an average of 41 word pairs per relation. A prototypicality score for each participant-generated word pair was calculated based on how often a second group of participants chose the word pair as the best and worst example of the relation among a set of choices. Table~\ref{relation-examples} shows examples illustrating two representative subtypes of each of the ten general types of relations.

\begin{figure*}[!ht]
\vspace{-5mm}
\centering
\includegraphics[trim={50mm 48mm 50mm 25mm},clip,width=0.90\linewidth,keepaspectratio]{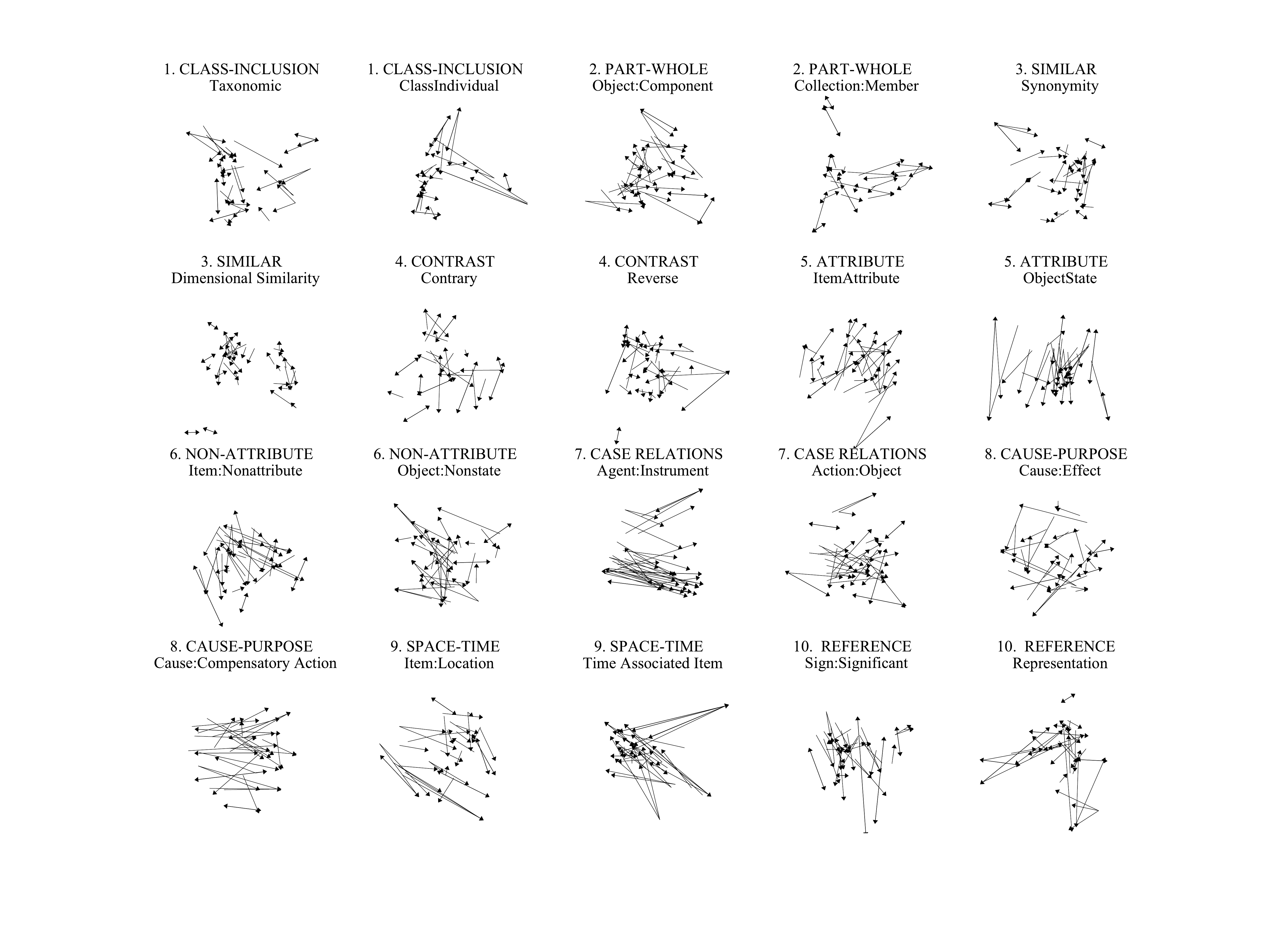}
\vspace{-4mm}
\caption{Visualizations of difference vectors for 20 relational subtypes using two-dimensional projections of \texttt{word2vec} word vectors obtained separately for each relation using principal components analysis.}
\label{projections}
\vspace{-4mm}
\end{figure*}

\setlength\tabcolsep{4pt}
\newcolumntype{P}[1]{>{\raggedright\arraybackslash}p{#1}}  
\begin{table}[tbh]
\caption{Examples of word pairs instantiating each of two representative subtypes from each general relation type in the SemEval-2012 Task 2 dataset}
\label{relation-examples}
\vspace{\baselineskip}
  \begin{tabular*}{0.48\textwidth}{lP{2.8cm}l}
  \toprule
  Relation type & Subtype & Example \\
  \midrule
  1. \textsc{class-}	& \textit{Taxonomic}			& \textit{flower} : \textit{tulip} \\
  \hspace{1em}\textsc{inclusion}	& \textit{Class:Individual}	& \textit{river} : \textit{Nile} \\
  \midrule
  \multirow{2}{*}{2. \textsc{part-whole}}	& \textit{Object:Component}			& \textit{car} : \textit{engine} \\
 						& \textit{Collection:Member}	& \textit{forest} : \textit{tree} \\
  \midrule
  \multirow{3}{*}{3. \textsc{similar}}	& \textit{Synonymy}			& \textit{car} : \textit{auto} \\
 						& \textit{Dimensional Simi-}	& \textit{simmer} : \textit{boil} \\
                        & \textit{larity} & \\
  \midrule
  \multirow{2}{*}{4. \textsc{contrast}}	& \textit{Contrary}			& \textit{old} : \textit{young} \\
 						& \textit{Reverse}	& \textit{buy} : \textit{sell} \\
  \midrule
  \multirow{2}{*}{5. \textsc{attribute}}	& \textit{Item:Attribute}			& \textit{beggar} : \textit{poor} \\
 						& \textit{Object:State}	& \textit{coward} : \textit{fear} \\
  \midrule
  \multirow{2}{*}{6. \textsc{non-attribute}} 		& \textit{Item:Nonattribute} & \textit{fire} : \textit{cold} \\
  						& \textit{Object:Nonstate}	& \textit{corpse} : \textit{life} \\
  \midrule
  7. \textsc{case} 		& \textit{Agent:Instrument} & \textit{soldier} : \textit{gun} \\
  \hspace{1em}\textsc{relations}	& \textit{Action:Object}	& \textit{plow} : \textit{earth} \\
  \midrule
  \multirow{3}{*}{8. \textsc{cause-purpose}} & \textit{Cause:Effect} & \textit{joke} : \textit{laughter} \\
  						& \textit{Cause:Compensa-}	& \textit{hunger} : \textit{eat} \\
                        & \textit{tory action} & \\
  \midrule
  \multirow{3}{*}{9. \textsc{space-time}} & \textit{Location:Item} & \textit{library} : \textit{book} \\
  						& \textit{Time:Associated}	& \textit{winter} : \textit{snow} \\
                        & \textit{Item} & \\
  \midrule
  \multirow{2}{*}{10. \textsc{reference}} & \textit{Sign:Significant} & \textit{siren} : \textit{danger} \\
  						& \textit{Representation}	& \textit{diary} : \textit{person} \\
  \bottomrule
  \end{tabular*}
\vspace{-4.9mm}
\end{table}
According to the parallelogram model, two pairs of words (\textit{A} : \textit{B} and \textit{C} : \textit{D}) are relationally similar to the extent that their difference vectors ($v_B - v_A$ and $v_D - v_C$) are similar. How appropriate is this geometric relationship for the various semantic relations? As a preliminary investigation of this question, we projected the 300-dimensional \texttt{word2vec} vectors into a two-dimensional space using principal components analysis separately for each relational subtype in the SemEval dataset, and visualized the difference vectors for the participant-generated word pairs from each relation. Figure~\ref{projections} shows the difference vectors for the 20 relational subtypes that are shown in Table~\ref{relation-examples}.

Examining the difference vectors for each relation shows that the parallelogram rule does not appear to capture all relations. \textsc{case relations} \textit{Agent:Instrument} (e.g., \textit{farmer} : \textit{tractor}) shows a nearly perfect correspondence with what we would expect under the parallelogram model, with all difference vectors aligning. However, many of the relations appear to have no clear geometric pattern. Nevertheless, simply looking at projections of the difference vectors is not sufficient to evaluate the power of geometric models of relational similarity to capture various relations, because information is lost in the projections. What is required is a detailed evaluation on judgments of relational similarity between word pairs within each relation.

%
%
Although the SemEval-2012 dataset contains prototypicality scores for the participant-generated word pairs within each relation, which have been interpreted as the relational similarities between the participant-generated pairs and the paradigmatic pairs, prototypicality is influenced by other factors such as the production frequencies of words \citep{Uyeda1980}. Moreover, because participants were encouraged to focus on the \textit{relation} illustrated by the paradigmatic examples, the prototypicality scores may not have much to do with the particular word pairs chosen as paradigmatic examples. Experiment 1 aims to address these problems.

%
%
%
\subsection{Experiment 1: Benchmarking Relational Similarity}
To overcome the limitations of the SemEval-2012 Task 2 dataset for our purposes, we collected a new large dataset that \textit{directly} measures human judgments of relational similarity between word pairs, focusing on comparisons between word pairs with similar relations.

\subsubsection{Participants}
We recruited 823 participants from Amazon Mechanical Turk. Participants were paid \$2.00 for the 20-minute study. We excluded 158 participants from the data analysis because they failed two or more of the attention checks (see below).

\subsubsection{Stimuli}
The stimuli for this study were taken from the SemEval-2012 Task 2 dataset. We were mainly interested in how people rate relational similarities between participant-generated word pairs within each of the 79 relational subtypes. However, because the total number of such ``within-subtype'' pairwise comparisons is still enormous, we selected the most representative subtype for each relation type out of the two shown in Table~\ref{relation-examples}. The subtype we chose is the first of each pair of subtypes that appears in Table~\ref{relation-examples}. We then randomly chose 30 word pairs out of the entire participant-generated set for each of the 10 subtypes and formed all possible within-subtype comparisons between these word pairs. This created a set of 4,350 within-subtype comparisons. Finally, in order to encourage participants to use the entire rating scale, we added 925 ``between-subtype'' comparisons, which are comparisons between word pairs from different subtypes within a type (e.g., \textit{Object:Component} and \textit{Collection:Member}, both subtypes of \textsc{part-whole}), and 925 ``between-type'' comparisons, which are comparisons between word pairs from the representative subtypes of different relational types (e.g., \textit{Object:Component} and \textit{Taxonomic \textsc{class-inclusion}}).
\subsubsection{Procedure}
Participants were given instructions about relational similarity, which included an example of two word pairs that have similar relationships (\textit{kitten} : \textit{cat} and \textit{chick} : \textit{chicken}) and an example of word pairs with dissimilar relationships (\textit{chick} : \textit{chicken} and \textit{hen} : \textit{rooster}). Participants then viewed two pairs of words side-by-side on each page and were asked to rate the similarity of the relationships shown by the two word pairs on a scale from 1 (extremely different) to 7 (extremely similar). They rated 100 comparisons in a random order, 70 of which were within-subtype, 15 of which were between-subtype, and 15 of which were between-type. The left-right order of the two word pairs on the screen was chosen randomly (but order within pairs was of course maintained). After every 20 trials, there was an attention check question that asked participants to indicate whether two words are the same or different.

\subsubsection{Results \& Discussion}
We obtained at least 10 good ratings for each comparison, with an average of 10.74 ratings per comparison. The mean rating across all comparisons was 4.52 (\textit{SD} = 2.17). As expected, we obtained the highest relational similarity ratings for within-subtype comparisons (\textit{M} = 5.01, \textit{SD} = 1.98), mid-level ratings for between-subtype comparisons (\textit{M} = 4.02, \textit{SD} = 2.14) and the lowest ratings for between-type comparisons (\textit{M} = 2.70, \textit{SD} = 1.93).

We calculated relational similarity for each comparison using \texttt{word2vec} and \texttt{GloVe} word representations. We used the 300-dimensional \texttt{word2vec} vectors trained on the Google News corpus that were provided by Google \citep{Mikolov2013-word2vec1}, and the 300-dimensional \texttt{GloVe} vectors trained on a Common Crawl web crawl corpus that were provided by \cite{pennington2014glove}. We tested two measures of similarity between difference vectors, cosine similarity and Euclidean distance. Specifically, for a given comparison between two word pairs, \textit{A} : \textit{B} and \textit{C} : \textit{D}, letting $\mathbf{r}_1 = v_B - v_A$ and $\mathbf{r}_2 = v_D - v_C$, we calculated the cosine similarity,
\begin{equation*}
\frac{\mathbf{r}_1 \cdot \mathbf{r}_2}{\lVert \mathbf{r}_1\rVert\lVert \mathbf{r}_2\rVert},
\end{equation*}
\noindent
as well as a similarity measure based on Euclidean distance,
\begin{equation*}
1 - \lVert \mathbf{r}_1 - \mathbf{r}_2 \rVert .
\end{equation*}
Cosine similarity is typically used to measure similarity in vector spaces such as \texttt{word2vec} and \texttt{GloVe}. However, using Euclidean distance corresponds more closely to the original parallelogram model, in which not only the directions but also the lengths of the difference vectors needed to be similar for two word pairs to be considered relationally similar.

\begin{figure*}
\centering
\includegraphics[trim = 0mm 1mm 0 0, clip, width=\linewidth,keepaspectratio]{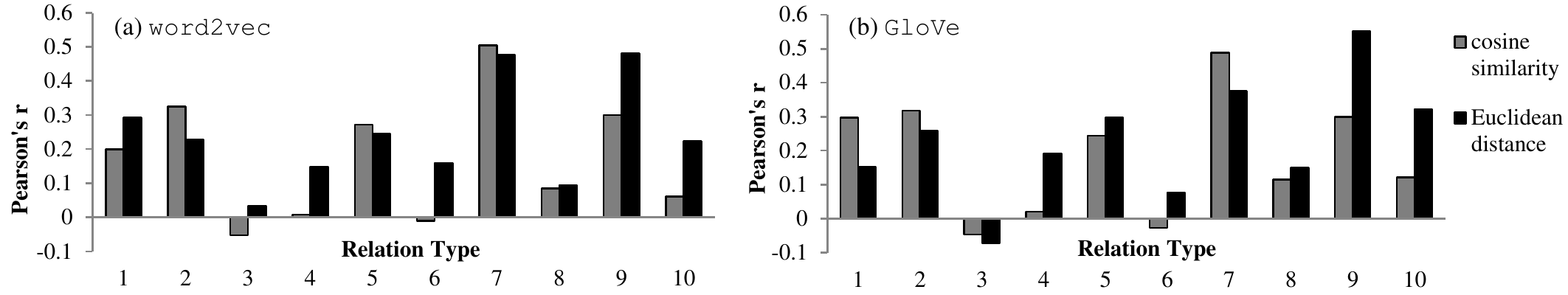}
\vspace{-6mm}
\caption{Pearons's $r$ between human relational similarity ratings and model predictions on different relation types for (a) \texttt{word2vec} and (b) \texttt{GloVe}. The name and examples of each numbered relation type are shown in Table~\ref{relation-examples}.}
\label{relsim-both}
\vspace{-4mm}
\end{figure*}

Figure \ref{relsim-both} shows Pearson's correlations between predicted relational similarity scores and average human relational similarity ratings on each relation type (including both within-subtype and between-subtype comparisons) for each vector space and similarity measure. There is considerable variation in the performance of \texttt{word2vec} and \texttt{GloVe} in predicting human relational similarity ratings. As might be expected from examining Figure \ref{projections}, cosine similarity performs the best on \textsc{case relations} (relation 7). However, cosine similarity completely fails on \textsc{similar} (relation 3), \textsc{contrast} (relation 4), and \textsc{non-attribute} (relation 6). Euclidean distance boosts performance on the latter two relations, but still under-performs overall compared to most other relations. Nevertheless, Euclidean distance does perform very well on \textsc{space-time} (relation 9).

%
%
These results indicate that a single relational comparison strategy cannot capture all semantic relations in the spaces provided. It is unclear if such a result is a reflection of the word embeddings or actual variation in human analogical strategies. Next, we turn to the broader question of the appropriateness of the class of geometric models in general for representing human relational similarity behavior.
%

\section{Violations of Geometric Constraints}
Distance metrics in vector spaces must obey certain geometric constraints, such as symmetry (the distance from \textit{x} to \textit{y} is the same as the distance from \textit{y} to \textit{x}) and the triangle inequality (if the distance between \textit{x} and \textit{y} is small and the distance between \textit{y} and \textit{z} is small, then the distance between \textit{x} and \textit{z} cannot be very large). Cosine similarity, used to measure similarity between \texttt{word2vec} representations, also obeys symmetry and an analogue of the triangle inequality \citep{griffiths2007topics}. However, psychological representations of similarity do not always obey these constraints \citep{tversky1977features}. The famous example of this is that people judge North Korea to be more similar to China than the other way around, a violation of symmetry. \cite{griffiths2007topics} examined the word representations derived by Latent Semantic Analysis \citep{landauer1997solution}, another well-known vector space model, and found that these representations are unable to account for violations of symmetry and the triangle inequality in human word association data. Nevertheless, all prior work has focused on first-order similarity between words, and second-order (relational) similarity between word pairs might be expected to follow a different pattern. In this section, we show that human judgments of relational similarity also do not satisfy the geometric constraints of symmetry and the triangle inequality. Vector space models such as \texttt{word2vec} and \texttt{GloVe} cannot account for these violations.

%
%
%
\subsection{Experiment 2: Symmetry}
In this experiment, we examined whether there were any pairs of word pairs for which participants' judgments of relational similarity changed when the presentation order was reversed. We might expect such asymmetry to occur when a word pair has multiple relations and shares ones of its less salient relations with another word pair. For example, when presented with \textit{angry} : \textit{smile} -- \textit{exhausted} : \textit{run}, one might think, ``an angry person doesn't want to smile" and ``an exhausted person doesn't want to run," but when presented with \textit{exhausted} : \textit{run} -- \textit{angry} : \textit{smile}, one might think,``running makes a person exhausted, but smiling doesn't make a person angry." Thus, participants might give high relational similarity ratings in the first presentation order and low ratings in the second order.\\\\
%
%
%
\textbf{Participants} \, We recruited 1,102 participants from Amazon Mechanical Turk, who gave informed consent and were paid \$1.00 for the 10-minute study. We excluded 99 participants from the data analysis because they failed two or more of the attention checks (see below).

\subsubsection{Stimuli}
We randomly selected 220 within-subtype, 220 between-subtype, and 60 between-type comparisons from all possible comparisons formed using the entire SemEval-2012 Task 2 dataset. We created two versions of each comparison, in which the order of the word pairs were switched.

\subsubsection{Procedure}
Participants were given instructions about relational similarity and the two examples used in Experiment 1 illustrating similar and dissimilar relationships. They saw one word pair in each comparison first and were asked to think of the relationship between the words. Then after a 600 ms delay, the other word pair was shown and participants were asked to rate the similarity of the relationships on a 7-point scale. Participants rated 50 comparisons, including 22 within-subtype, 22 between-subtype, and 6 between-type comparisons. Each participant viewed each comparison in only one of its presentation orders. After every 10 trials, there was an attention check question that asked participants to indicate whether two words are the same or different.

\subsubsection{Results \& Discussion}
We obtained about 50 ratings for each comparison in each presentation order. We conducted a \textit{t}-test for each comparison to see if the two presentation orders resulted in significantly different relational similarity ratings. 77 of these \textit{t}-tests were statistically significant at the .05 level. The number of \textit{t}-tests that we would expect to be significant at the $\alpha = 0.05$ level if presentation order did not matter for any of the comparisons is 25. Assuming that the \textit{t}-tests are independent, a binomial test reveals that this deviation is statistically significant, $p < .001$.

Examining the comparisons for which different presentation orders resulted in significantly different relational similarity ratings confirms our guess as to when people's judgments of relational similarity might not obey symmetry. The previously mentioned example of \textit{angry} : \textit{smile} and \textit{exhausted} : \textit{run} indeed elicited higher ratings in the direction shown here (4.76 mean rating) than in the opposite direction (2.36 mean rating). As another example, people rated \textit{hairdresser} : \textit{comb} -- \textit{pitcher} : \textit{baseball} as more relationally similar (6.10 mean rating) than \textit{pitcher} : \textit{baseball} -- \textit{hairdresser} : \textit{comb} (4.84 mean rating). In the first presentation order, participants might be thinking that ``a hairdresser handles a comb'' and ``a pitcher handles a baseball,'' whereas in the second presentation order, they might be thinking ``a pitcher plays a specific role in baseball,'' which doesn't fit with \textit{hairdresser} : \textit{comb}.


\subsection{Experiment 3: Triangle Inequality}
For this experiment, we created triads of word pairs for which we expected people's relational similarity judgments to violate the triangle inequality, such as \textit{nurse} : \textit{patient}, \textit{mother} : \textit{baby}, and \textit{frog} : \textit{tadpole}. This triad violates the triangle inequality because \textit{nurse} : \textit{patient} :: \textit{mother} : \textit{baby} is a good analogy (relationally similar), and so is \textit{mother} : \textit{baby} :: \textit{frog} : \textit{tadpole}, but \textit{nurse} : \textit{patient} :: \textit{frog} : \textit{tadpole} is not. In this example, the middle pair has multiple relations and shares one of them with the first pair and a different one with the last pair. We presented the two word pairs in each analogy together and asked participants to rate the quality of the analogy rather than relational similarity, because we wanted to encourage participants to consider the two relations together rather than using one relation as a reference.

\subsubsection{Participants}
We recruited 71 participants from Amazon Mechanical Turk, who gave informed consent and were paid \$0.50 for the 5-minute study. This group of participants did not overlap with the participants in Experiment 2. We excluded 11 participants from the data analysis because they failed one of the attention checks (see below).

\subsubsection{Stimuli}
We created twelve triads of word pairs for which analogy quality judgments are likely to violate the triangle inequality. For every triad, the analogy formed between the first and third word pairs was expected to be rated low and the other two analogies were expected to be rated highly.

\subsubsection{Procedure}
Participants were given instructions about verbal analogies and the two examples used in Experiments 1 and 2 as examples of good and bad analogies, respectively. They were then asked to rate the quality of each analogy on a scale from 1 (very bad) to 7 (very good). For each of the twelve triads, each participant viewed one of the three analogies. Each participant received four analogies formed between the first and second word pairs of various triads (analogy type 1-2), four formed between the second and third word pairs (type 2-3), and four formed between the first and third word pairs (type 1-3). Because two thirds of these analogies are expected to be rated highly, participants also viewed four ``filler'' analogies expected to be given low ratings. Finally, there were two attention check questions that asked to participants to simply choose 1 (or 7) for a bad (or good) analogy.

\subsubsection{Results \& Discussion}
We obtained 20 ratings for each analogy. We calculated the mean participant rating for each analogy and conducted a one-way between-subjects ANOVA to test if there was an effect of the analogy type (1-2, 2-3, or 1-3) on the mean analogy quality rating. This revealed a significant effect of analogy type, $F(2, 33) = 45.57$, $p < 0.001$. Post hoc comparisons using the Tukey HSD test indicated that the mean ratings for both type 1-2 analogies ($M = 5.44$, $SD = .99$) and type 2-3 analogies ($M = 5.43$, $SD = .63$) were significantly higher than the mean rating for type 1-3 analogies ($M = 2.99$, $SD = .46$), $p < .001$, whereas the mean ratings for type 1-2 and type 2-3 analogies did not differ significantly from each other. This is consistent with our expectation that types 1-2 and 2-3 analogies would both be rated highly, whereas type 1-3 analogies would be rated lowly. These results indicate that participants' analogy quality ratings violated the triangle inequality.

We obtained the predicted relational similarity between the word pairs in each analogy by calculating the cosine similarity between difference vectors using \texttt{word2vec} and \texttt{GloVe} representations. Then we conducted separate ANOVAs for the two representational spaces to test whether there was an effect of analogy type on the predicted relational similarity in each space. Neither ANOVA indicated a significant effect of analogy type. For \texttt{word2vec}, $F(2, 33) = 1.20$, $p = .31$. For \texttt{GloVe}, $F(2, 33) = .24$, $p = .79$. These results indicate that the predictions of relational similarity made by \texttt{word2vec} and \texttt{GloVe} do not violate the triangle inequality for our stimuli.

\begin{figure}[!t]
\centering
\includegraphics[trim = 3mm 0mm 0mm 3mm, clip, width=0.9\linewidth,keepaspectratio]{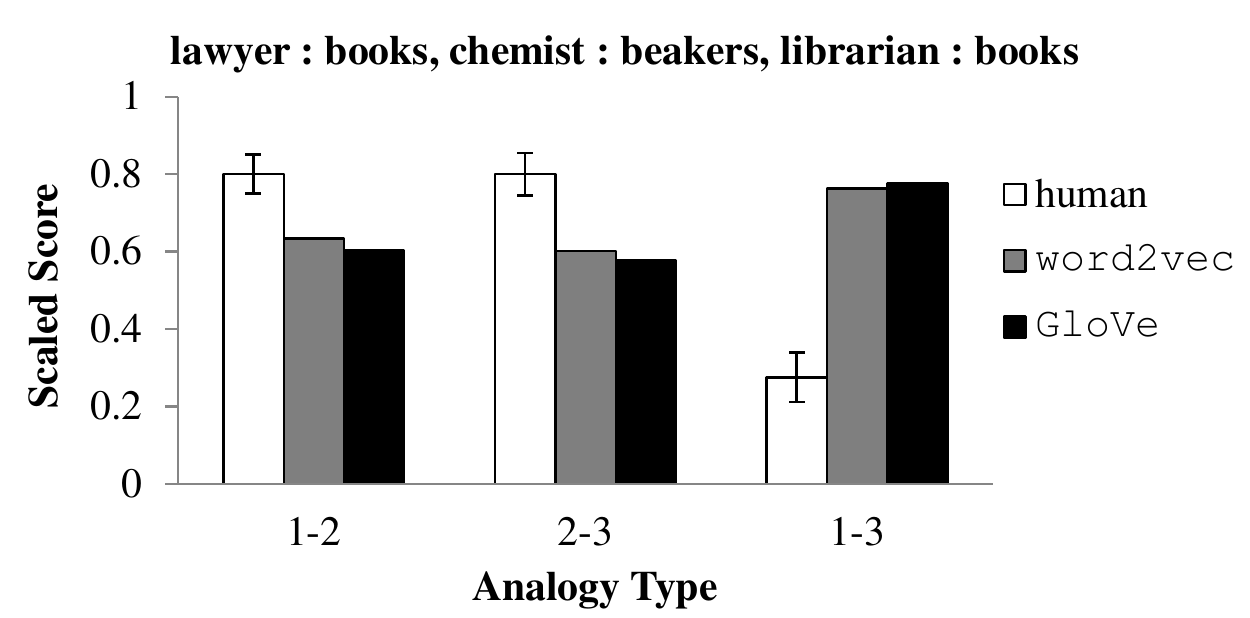}
\vspace{-4mm}
\caption{Mean human ratings and predicted relational similarities (scaled to the range 0-1) for the triad \textit{lawyer} : \textit{books}, \textit{chemist} : \textit{beakers}, and \textit{librarian} : \textit{books}. Error bars indicate 1 SEM for scaled human ratings.}
\label{triangle-inquality-example}
\vspace{-5mm}
\end{figure}

Because each participant contributed ratings to more than one analogy (although only one per triad), the observations are not entirely independent in the first overall ANOVA that we conducted on the participant data. Thus, we conducted a separate between-subjects ANOVA for each of the twelve triads to test if there was an effect of analogy type (1-2, 2-3, or 1-3) on the analogy quality ratings for each triad. All twelve ANOVAs were significant, with every $p < .01$. We then conducted post hoc comparisons using the Tukey HSD test. The pattern we expected was that types 1-2 and 2-3 analogies would have significantly higher ratings than type 1-3 analogies, but would not differ significantly from each other. We observed this pattern for seven of the twelve triads. For every one of the remaining triads, the mean ratings for the type 1-2 and type 2-3 analogies were higher than the mean rating for the type 2-3 analogy, but which differences were statistically significant differed among the triads. Figure~\ref{triangle-inquality-example} shows an example of one of the seven triads with the expected pattern and compares the mean participant ratings and predicted relational similarities for the three analogies.
\vspace{-1mm}
\section{Conclusions}
Our results provide a clearer picture of the utility of vector-space models of analogy. The parallelogram model makes good predictions of human relational similarity judgments for some relations, but is less appropriate for others. For example, consider the word pairs represented as vectors in Figure \ref{projections}. As one would expect, relation \textsc{similar} seems to be best represented by a short difference vector rather than the direction of the difference vector. More generally, in more complex analogies with the source and target each consisting of many points in the vector space, one could imagine many ways of describing relationships between the two sets of points.


More challenging are the constraints posed by the geometric axioms. In our datasets, we found considerable violations of two of these axioms, which cannot be overcome through better embedding methods. In light of this, it would be interesting to follow the history of models of first-order similarity in considering the use of featural representations \citep{tversky1977features}, exploring methods of measuring similarity in vector spaces that are no longer subject to the constraints imposed by the metric axioms \citep{krumhansl78}, or reformulating the problem as probabilistic inference \citep{griffiths2007topics}.\

\smallskip
\smallskip

\noindent{\bf Acknowledgments.} Supported by grant number FA9550-13-1-0170 from the Air Force Office of Scientific Research.
\vspace{-2mm}
\renewcommand{\bibliographytypesize}{\small}
\bibliographystyle{apacite}

\bibliography{CogSci_Template}

\begin{thebibliography}{}

\bibitem [\protect \citeauthoryear {%
Bejar%
, Chaffin%
\BCBL {}\ \BBA {} Embretson%
}{%
Bejar%
\ \protect \BOthers {.}}{%
{\protect \APACyear {1991}}%
}]{%
Bejar1991}
\APACinsertmetastar {%
Bejar1991}%
\begin{APACrefauthors}%
Bejar, I\BPBI I.%
, Chaffin, R.%
\BCBL {}\ \BBA {} Embretson, S.%
\end{APACrefauthors}%
\unskip\
\newblock
\APACrefYearMonthDay{1991}{}{}.
\newblock
{\BBOQ}\APACrefatitle {A taxonomy of semantic relations} {A taxonomy of
  semantic relations}.{\BBCQ}
\newblock
\BIn{} \APACrefbtitle {Cognitive and psychometric analysis of analogical
  problem solving} {Cognitive and psychometric analysis of analogical problem
  solving}\ (\BPGS\ 55--91).
\newblock
\APACaddressPublisher{}{Springer}.
\PrintBackRefs{\CurrentBib}

\bibitem [\protect \citeauthoryear {%
Doumas%
, Hummel%
\BCBL {}\ \BBA {} Sandhofer%
}{%
Doumas%
\ \protect \BOthers {.}}{%
{\protect \APACyear {2008}}%
}]{%
doumas2008theory}
\APACinsertmetastar {%
doumas2008theory}%
\begin{APACrefauthors}%
Doumas, L\BPBI A.%
, Hummel, J\BPBI E.%
\BCBL {}\ \BBA {} Sandhofer, C\BPBI M.%
\end{APACrefauthors}%
\unskip\
\newblock
\APACrefYearMonthDay{2008}{}{}.
\newblock
{\BBOQ}\APACrefatitle {A theory of the discovery and predication of relational
  concepts.} {A theory of the discovery and predication of relational
  concepts.}{\BBCQ}
\newblock
\APACjournalVolNumPages{Psychological review}{115}{1}{1}.
\PrintBackRefs{\CurrentBib}

\bibitem [\protect \citeauthoryear {%
Ehresman%
\ \BBA {} Wessel%
}{%
Ehresman%
\ \BBA {} Wessel%
}{%
{\protect \APACyear {1978}}%
}]{%
ehresman1978perception}
\APACinsertmetastar {%
ehresman1978perception}%
\begin{APACrefauthors}%
Ehresman, D.%
\BCBT {}\ \BBA {} Wessel, D\BPBI L.%
\end{APACrefauthors}%
\unskip\
\newblock
\APACrefYear{1978}.
\newblock
\APACrefbtitle {Perception of timbral analogies} {Perception of timbral
  analogies}.
\newblock
\APACaddressPublisher{}{Centre Georges Pompidou}.
\PrintBackRefs{\CurrentBib}

\bibitem [\protect \citeauthoryear {%
Gentner%
\ \BBA {} Forbus%
}{%
Gentner%
\ \BBA {} Forbus%
}{%
{\protect \APACyear {2011}}%
}]{%
gentner2011computational}
\APACinsertmetastar {%
gentner2011computational}%
\begin{APACrefauthors}%
Gentner, D.%
\BCBT {}\ \BBA {} Forbus, K\BPBI D.%
\end{APACrefauthors}%
\unskip\
\newblock
\APACrefYearMonthDay{2011}{}{}.
\newblock
{\BBOQ}\APACrefatitle {Computational models of analogy} {Computational models
  of analogy}.{\BBCQ}
\newblock
\APACjournalVolNumPages{Wiley interdisciplinary reviews: cognitive
  science}{2}{3}{266--276}.
\PrintBackRefs{\CurrentBib}

\bibitem [\protect \citeauthoryear {%
Griffiths%
, Steyvers%
\BCBL {}\ \BBA {} Tenenbaum%
}{%
Griffiths%
\ \protect \BOthers {.}}{%
{\protect \APACyear {2007}}%
}]{%
griffiths2007topics}
\APACinsertmetastar {%
griffiths2007topics}%
\begin{APACrefauthors}%
Griffiths, T\BPBI L.%
, Steyvers, M.%
\BCBL {}\ \BBA {} Tenenbaum, J\BPBI B.%
\end{APACrefauthors}%
\unskip\
\newblock
\APACrefYearMonthDay{2007}{}{}.
\newblock
{\BBOQ}\APACrefatitle {Topics in semantic representation.} {Topics in semantic
  representation.}{\BBCQ}
\newblock
\APACjournalVolNumPages{Psychological Review}{114}{2}{211}.
\PrintBackRefs{\CurrentBib}

\bibitem [\protect \citeauthoryear {%
Jurgens%
, Turney%
, Mohammad%
\BCBL {}\ \BBA {} Holyoak%
}{%
Jurgens%
\ \protect \BOthers {.}}{%
{\protect \APACyear {2012}}%
}]{%
Jurgens2012}
\APACinsertmetastar {%
Jurgens2012}%
\begin{APACrefauthors}%
Jurgens, D\BPBI A.%
, Turney, P\BPBI D.%
, Mohammad, S\BPBI M.%
\BCBL {}\ \BBA {} Holyoak, K\BPBI J.%
\end{APACrefauthors}%
\unskip\
\newblock
\APACrefYearMonthDay{2012}{}{}.
\newblock
{\BBOQ}\APACrefatitle {Semeval-2012 task 2: Measuring degrees of relational
  similarity} {Semeval-2012 task 2: Measuring degrees of relational
  similarity}.{\BBCQ}
\newblock
\BIn{} \APACrefbtitle {Proceedings of the First Joint Conference on Lexical and
  Computational Semantics-Volume 1} {Proceedings of the first joint conference
  on lexical and computational semantics-volume 1}\ (\BPGS\ 356--364).
\PrintBackRefs{\CurrentBib}

\bibitem [\protect \citeauthoryear {%
Krumhansl%
}{%
Krumhansl%
}{%
{\protect \APACyear {1978}}%
}]{%
krumhansl78}
\APACinsertmetastar {%
krumhansl78}%
\begin{APACrefauthors}%
Krumhansl, C.%
\end{APACrefauthors}%
\unskip\
\newblock
\APACrefYearMonthDay{1978}{}{}.
\newblock
{\BBOQ}\APACrefatitle {Concerning the applicability of geometric models to
  similarity data: The interrelationship between similarity and spatial
  density} {Concerning the applicability of geometric models to similarity
  data: The interrelationship between similarity and spatial density}.{\BBCQ}
\newblock
\APACjournalVolNumPages{Psychological Review}{85}{}{450-463}.
\PrintBackRefs{\CurrentBib}

\bibitem [\protect \citeauthoryear {%
Landauer%
\ \BBA {} Dumais%
}{%
Landauer%
\ \BBA {} Dumais%
}{%
{\protect \APACyear {1997}}%
}]{%
landauer1997solution}
\APACinsertmetastar {%
landauer1997solution}%
\begin{APACrefauthors}%
Landauer, T\BPBI K.%
\BCBT {}\ \BBA {} Dumais, S\BPBI T.%
\end{APACrefauthors}%
\unskip\
\newblock
\APACrefYearMonthDay{1997}{}{}.
\newblock
{\BBOQ}\APACrefatitle {A solution to Plato's problem: The latent semantic
  analysis theory of acquisition, induction, and representation of knowledge.}
  {A solution to plato's problem: The latent semantic analysis theory of
  acquisition, induction, and representation of knowledge.}{\BBCQ}
\newblock
\APACjournalVolNumPages{Psychological Review}{104}{2}{211}.
\PrintBackRefs{\CurrentBib}

\bibitem [\protect \citeauthoryear {%
Lu%
, Chen%
\BCBL {}\ \BBA {} Holyoak%
}{%
Lu%
\ \protect \BOthers {.}}{%
{\protect \APACyear {2012}}%
}]{%
lu2012bayesian}
\APACinsertmetastar {%
lu2012bayesian}%
\begin{APACrefauthors}%
Lu, H.%
, Chen, D.%
\BCBL {}\ \BBA {} Holyoak, K\BPBI J.%
\end{APACrefauthors}%
\unskip\
\newblock
\APACrefYearMonthDay{2012}{}{}.
\newblock
{\BBOQ}\APACrefatitle {Bayesian analogy with relational transformations.}
  {Bayesian analogy with relational transformations.}{\BBCQ}
\newblock
\APACjournalVolNumPages{Psychological review}{119}{3}{617}.
\PrintBackRefs{\CurrentBib}

\bibitem [\protect \citeauthoryear {%
Mikolov%
, Sutskever%
, Chen%
, Corrado%
\BCBL {}\ \BBA {} Dean%
}{%
Mikolov%
\ \protect \BOthers {.}}{%
{\protect \APACyear {2013}}%
}]{%
Mikolov2013-word2vec1}
\APACinsertmetastar {%
Mikolov2013-word2vec1}%
\begin{APACrefauthors}%
Mikolov, T.%
, Sutskever, I.%
, Chen, K.%
, Corrado, G\BPBI S.%
\BCBL {}\ \BBA {} Dean, J.%
\end{APACrefauthors}%
\unskip\
\newblock
\APACrefYearMonthDay{2013}{}{}.
\newblock
{\BBOQ}\APACrefatitle {Distributed representations of words and phrases and
  their compositionality} {Distributed representations of words and phrases and
  their compositionality}.{\BBCQ}
\newblock
\BIn{} \APACrefbtitle {Advances in neural information processing systems}
  {Advances in neural information processing systems}\ (\BPGS\ 3111--3119).
\PrintBackRefs{\CurrentBib}

\bibitem [\protect \citeauthoryear {%
Pennington%
, Socher%
\BCBL {}\ \BBA {} Manning%
}{%
Pennington%
\ \protect \BOthers {.}}{%
{\protect \APACyear {2014}}%
}]{%
pennington2014glove}
\APACinsertmetastar {%
pennington2014glove}%
\begin{APACrefauthors}%
Pennington, J.%
, Socher, R.%
\BCBL {}\ \BBA {} Manning, C\BPBI D.%
\end{APACrefauthors}%
\unskip\
\newblock
\APACrefYearMonthDay{2014}{}{}.
\newblock
{\BBOQ}\APACrefatitle {Glove: Global Vectors for Word Representation.} {Glove:
  Global vectors for word representation.}{\BBCQ}
\newblock
\BIn{} \APACrefbtitle {EMNLP} {Emnlp}\ (\BVOL~14, \BPGS\ 1532--1543).
\PrintBackRefs{\CurrentBib}

\bibitem [\protect \citeauthoryear {%
Radford%
, Metz%
\BCBL {}\ \BBA {} Chintala%
}{%
Radford%
\ \protect \BOthers {.}}{%
{\protect \APACyear {2015}}%
}]{%
radford2015unsupervised}
\APACinsertmetastar {%
radford2015unsupervised}%
\begin{APACrefauthors}%
Radford, A.%
, Metz, L.%
\BCBL {}\ \BBA {} Chintala, S.%
\end{APACrefauthors}%
\unskip\
\newblock
\APACrefYearMonthDay{2015}{}{}.
\newblock
{\BBOQ}\APACrefatitle {Unsupervised Representation Learning with Deep
  Convolutional Generative Adversarial Networks} {Unsupervised representation
  learning with deep convolutional generative adversarial networks}.{\BBCQ}
\newblock
\APACjournalVolNumPages{arXiv preprint arXiv:1511.06434}{}{}{}.
\PrintBackRefs{\CurrentBib}

\bibitem [\protect \citeauthoryear {%
Reed%
, Zhang%
, Zhang%
\BCBL {}\ \BBA {} Lee%
}{%
Reed%
\ \protect \BOthers {.}}{%
{\protect \APACyear {2015}}%
}]{%
reed2015deep}
\APACinsertmetastar {%
reed2015deep}%
\begin{APACrefauthors}%
Reed, S\BPBI E.%
, Zhang, Y.%
, Zhang, Y.%
\BCBL {}\ \BBA {} Lee, H.%
\end{APACrefauthors}%
\unskip\
\newblock
\APACrefYearMonthDay{2015}{}{}.
\newblock
{\BBOQ}\APACrefatitle {Deep Visual Analogy-Making} {Deep visual
  analogy-making}.{\BBCQ}
\newblock
\BIn{} \APACrefbtitle {Advances in Neural Information Processing Systems}
  {Advances in neural information processing systems}\ (\BPGS\ 1252--1260).
\PrintBackRefs{\CurrentBib}

\bibitem [\protect \citeauthoryear {%
Richland%
\ \BBA {} Simms%
}{%
Richland%
\ \BBA {} Simms%
}{%
{\protect \APACyear {2015}}%
}]{%
richland2015analogy}
\APACinsertmetastar {%
richland2015analogy}%
\begin{APACrefauthors}%
Richland, L\BPBI E.%
\BCBT {}\ \BBA {} Simms, N.%
\end{APACrefauthors}%
\unskip\
\newblock
\APACrefYearMonthDay{2015}{}{}.
\newblock
{\BBOQ}\APACrefatitle {Analogy, higher order thinking, and education} {Analogy,
  higher order thinking, and education}.{\BBCQ}
\newblock
\APACjournalVolNumPages{Wiley Interdisciplinary Reviews: Cognitive
  Science}{6}{2}{177--192}.
\PrintBackRefs{\CurrentBib}

\bibitem [\protect \citeauthoryear {%
Rumelhart%
\ \BBA {} Abrahamson%
}{%
Rumelhart%
\ \BBA {} Abrahamson%
}{%
{\protect \APACyear {1973}}%
}]{%
Rumelhart1973}
\APACinsertmetastar {%
Rumelhart1973}%
\begin{APACrefauthors}%
Rumelhart, D\BPBI E.%
\BCBT {}\ \BBA {} Abrahamson, A\BPBI A.%
\end{APACrefauthors}%
\unskip\
\newblock
\APACrefYearMonthDay{1973}{}{}.
\newblock
{\BBOQ}\APACrefatitle {A model for analogical reasoning} {A model for
  analogical reasoning}.{\BBCQ}
\newblock
\APACjournalVolNumPages{Cognitive Psychology}{5}{1}{1--28}.
\PrintBackRefs{\CurrentBib}

\bibitem [\protect \citeauthoryear {%
Tversky%
}{%
Tversky%
}{%
{\protect \APACyear {1977}}%
}]{%
tversky1977features}
\APACinsertmetastar {%
tversky1977features}%
\begin{APACrefauthors}%
Tversky, A.%
\end{APACrefauthors}%
\unskip\
\newblock
\APACrefYearMonthDay{1977}{}{}.
\newblock
{\BBOQ}\APACrefatitle {Features of similarity.} {Features of
  similarity.}{\BBCQ}
\newblock
\APACjournalVolNumPages{Psychological review}{84}{4}{327}.
\PrintBackRefs{\CurrentBib}

\bibitem [\protect \citeauthoryear {%
Uyeda%
\ \BBA {} Mandler%
}{%
Uyeda%
\ \BBA {} Mandler%
}{%
{\protect \APACyear {1980}}%
}]{%
Uyeda1980}
\APACinsertmetastar {%
Uyeda1980}%
\begin{APACrefauthors}%
Uyeda, K\BPBI M.%
\BCBT {}\ \BBA {} Mandler, G.%
\end{APACrefauthors}%
\unskip\
\newblock
\APACrefYearMonthDay{1980}{}{}.
\newblock
{\BBOQ}\APACrefatitle {Prototypicality norms for 28 semantic categories}
  {Prototypicality norms for 28 semantic categories}.{\BBCQ}
\newblock
\APACjournalVolNumPages{Behavior Research Methods \&
  Instrumentation}{12}{6}{587--595}.
\PrintBackRefs{\CurrentBib}

\end{thebibliography}

\end{document}